# Comparison of algorithms in Foreign Exchange Rate Prediction

Swagat Ranjit [1], Shruti Shrestha [2], Sital Subedi [3]
Department of Electronics and Computer Engineering
Advanced College of Engineering and Management
Tribhuvan University, Lalitpur, Nepal
Email: (swag.ranjit[1], srutsth[2], sitalsubedi777[3])@gmail.com

Subarna Shakya [4]
Department of Electronics and Computer Engineering
Pulchowk Campus, IOE
Tribhuvan University, Lalitpur, Nepal
Email: drss@ioe.edu.np

*Abstract*— Foreign currency exchange plays a vital role for trading of currency in the financial market. Due to its volatile nature, prediction of foreign currency exchange is a challenging task. This paper presents different machine learning techniques like Artificial Neural Network (ANN), Recurrent Neural Network (RNN) to develop prediction model between Nepalese Rupees against three major currencies Euro, Pound Sterling and US dollar. Recurrent Neural Network is a type of neural network that have feedback connections. In this paper, prediction model were based on different RNN architectures, feed forward ANN with back propagation algorithm and then compared the accuracy of each model. Different ANN architecture models like Feed forward neural network, Simple Recurrent Neural Network (SRNN), Gated Recurrent Unit (GRU) and Long Short Term Memory (LSTM) were used. Input parameters were open, low, high and closing prices for each currency. From this study, we have found that LSTM networks provided better results than SRNN and GRU networks.

I.    INTRODUCTION

It is a system of exchanging of currencies making the trade easier. It can be quoted as USD/NPR, EUR/NPR and GBP/NPR meaning that how much Nepalese Rupee can be paid to get equivalent currencies. The quotation USD/NPR 115.2 means that one American dollar is exchanged for 115.2 Nepalese Rupees. Trading at the right time with the correct approach can gain profit but a trade without knowledge can cause heavy loss. Hence, prediction of the financial market movement have become a challenging task [1]. One of the most fundamental concept of economics is the rule of supply and demand. Foreign exchange rate also works on the same principle. Currency values tend to rise with the increase in demand and tend to fall with the decrease in demand [2]. Factors such as stock market, political situation, economic news and interest rates affect the demand for currency change.

In previous days, different statistical models such as integrated moving average, auto linear regression were used for financial time series predictions. However, with the improvement in computing and processing, machine learning algorithms have made time series predictions possible. Neural Network is a part of Artificial Intelligence. ANN is an artificial model based on human brain that try to replicate the learning process of human brain. They have the power to analyze and predict. It can learn from a few examples or given inputs to learn and generate rules or operation [3]. ANN are applicable in different fields such as pattern recognition, stock market prediction, medical diagnoses, image processing, etc. [4]. RNN is a neural network that consist of feedback connections.

In this paper, we apply ANN for predicting foreign exchange rates of Nepalese Rupee (NPR) against three other currencies like American Dollar (USD), Pound Sterling (GBP) and EURO using their historical data. A total of 1500 historical exchange rates data for each of three currency rates were collected from Investing.com website and used as inputs to build the prediction model. Different ANN architectures were used to measure the performance of each currency.

Section II of this paper discusses about related works of this paper. Section III discusses the theory background methodology part. Section IV discusses about the results and analysis and section V discusses about the conclusion of paper.

II.    LITERATURE REVIEW

Kadilar & Adla [5] analyzed about Autoregressive Integrated Moving Average and different neural network models so as to predict Turkish Lira against American dollar. Results showed that neural network models produce better result, as compared to the ARIMA model, which has proved that ANN is better than ARIMA for financial time series prediction. (S. Kumar Chandar, et al., 2016) [6] explored on feed forward neural network so as to forecast American dollar against Indian Rupees. Results showed that accurate model was 1-4-1 with mean absolute percentage error 0.071. (Naeini et al., 2010) [7] worked on the feed forward neural network in which they used the historical data of the stock market in order to forecast the stock value. Results showed that MLP has lower accuracy in comparison with Elman RNN. Kamruzzaman et al., [8] used different Artificial Neural Network prediction training methods like scaled conjugate gradient and backpropagation to forecast American dollar and Euro currencies against Australian dollar.

It is found that most of them used the feed forward network with back propagation algorithm but it is too slow and tedious in real life scenario. Because of these cons, different neural network architecture like Deep Neural

Networks, Recurrent Neural Network, etc. are implemented in foreign exchange rate prediction[9]. (Samarawickrama et al., 2017) [10] compared performance of different ANN for predicting daily stock prices of Sri Lankan Stock Market. They built multi layer perceptron with Levenberg-Marquardt Algorithm, LSTM and GRU architectures for prediction of stock market. Different architectures were built by varying the number of hidden neurons of ANN and best result structure was chosen. Data was normalized in the range of [-1,1]. Results indicated that LSTM have better accuracy than other architecture of ANN.

Multilayer Perceptron (MLP) neural network consumes more time and not being able to remember or use the past data. To improve the performance of prediction models, Wei & Cheng [11] suggested a hybrid model that improves the Elman RNN model for forecasting Taiwan stock price.

III. THEORETICAL BACKGROUND

### A. Neural Network Model

There are two neural network model: Feed forward and Feedback neural networks. Feed-forward Multi Layer Perceptron network(Fig. 1) is the most widely used ANN trained by back propagation algorithm and data traverses from input to output layer and vice versa until minimum error is reached.

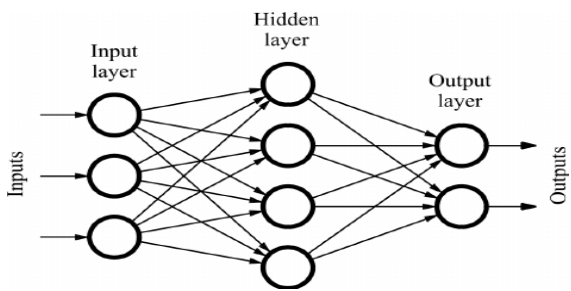

1. Feedforward Neural Network.

### B. Recurrent Neural Network

RNN is a feedback connection given to feedforward networks as explained by Hopfield in 1983 (Fig. 2). RNN is unique from other neural networks as it consists of internal memory to store the inputs received at each node. When taking decision, it considers the current as well as previously received inputs therefore, making the results more precise.

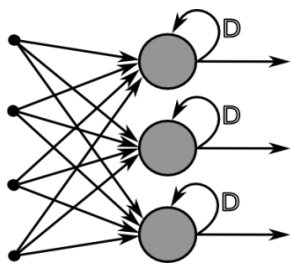

2. Feedforward Neural Network.

### C. Simple Recurrent Neural Network

A network having three layers arranged horizontally and including a set of context units is a special RNN model known as Simple Recurrent Neural Network[12]. This was first introduced by Jeff Elman. The weights are fixed in hidden layer which is connected to context layers. It can remember the past values of hidden units in context layer. Having this approach, it gives a better performance than feedforward network for sequential data prediction. The output layer is connected to neurons of context layers.

### D. Long Short Term Memory

Generally, RNN has a short term memory but in addition to LSTM, it can have a long term memory. This was introduced in the mid-90s so as to solve 'vanishing gradient problem'. It is a type of RNN that preserves the back propagated errors and those errors can be propagated backwards through layers. It consist of input gate, output gate and forget gate [13].

### E. Gated Recurrent Unit

GRU is a type of RNN which was introduced in 2014. It consist of a gating mechanism. It is similar to LSTM but it doesn't contain any output gate [14].

IV. METHODOLOGY

### A. Data Collection

Historical data are used for making a model and predicting the foreign currency exchange. Considering the complexity and training time of model, high, open, close and low price were selected as input because they were more correlated to output closing price. A large number of these sets of data were collected by crawling the Investing.com website over ten-months period.

The dataset are as follows (p-Current price):

- Yesterday's Low price (p-1)
- Yesterday's Open price (p-1)
- Yesterday's High price (p-1)
- Yesterday's Close price (p-1)

The output is taken as close price (p)

### B. Data Preprocessing

In this study, dataset were normalized between 0 to 1. It helps in reducing time and space complexity during processing. It is calculated using the following formula:

Normalized value = (Actual value – Minimum value) / (Maximum value – Minimum value)

### C. Neural Network Model

In this study, four different ANN architectures were used namely feed forward Neural Network trained with backpropagation algorithm, SRNN, LSTM and GRU.

For each model, 10 different structure of each architecture were built by changing the number of hidden units from 2 to 10. The number of input neuron is 4, with 2 to 10 varying hidden neurons and a single output neuron. The table I shows number of neuron in each model.

## D. Training And Testing Model

After normalization, the dataset were divided into training set, which consists of 70% of whole data, 15% for validation and the testing set consists of the rest 15% of the data.

Keras deep learning library was used to make RNN model in python on Linux platform. The activation functions and parameters were used to make the model as shown in table II.

Training was done so as to find the final synapse weight of each neuron. Each model was trained for 1500 iterations. The predicted value was for testing purpose using test dataset. Prediction was done to predict value for next day. To compare predicted and actual data, denormalization was done to convert them to actual format and then the predicted value were compared with actual data. To measure the performance of the models, Mean Absolute Error (MAE) was calculated so as to find the accuracy of the model. MAE value is inversely proportional to accuracy. So, the model with lowest MAE was selected as accurate model.

$$MAE = \sum_{i=1}^{n} \frac{|y_i - x_i|}{n}$$

where y = predicted value, x = actual value, n = number of data

I. NUMBER OF NEURON IN EACH MODEL

| Model No | Number of input neuron | Number of hidden neuron | Number of output neuron |
|---|---|---|---|
| 1 | 4 | 1 | 1 |
| 2 | 4 | 2 | 1 |
| 3 | 4 | 3 | 1 |
| 4 | 4 | 4 | 1 |
| 5 | 4 | 5 | 1 |
| 6 | 4 | 6 | 1 |
| 7 | 4 | 7 | 1 |
| 8 | 4 | 8 | 1 |
| 9 | 4 | 9 | 1 |

II. PARAMETER OF EACH MODEL

| Parameter | MLP | SRNN | LSTU | GRU |
|---|---|---|---|---|
| Activation | Sigmoid | Tan h | Tan h | Tan h |
| Loss function | MAE | MAE | MAE | MAE |
| Training algorithm | Back propagation | rmsprop | rmsprop | rmsprop |

## V. RESULT AND ANALYSIS

Table III, IV, V shows the result for different models of USD/NPR, EUR/NPR and GBP/NPR respectively along with their calculated MAE from the predicted results respectively.

III. BEST NETWORK MODEL FOR USD/NPR

| Model | Best Structure | MAE |
|---|---|---|
| MLP | 4-6-1 | 0.0858 |
| SRNN | 4-4-1 | 0.019 |
| GRU | 4-7-1 | 0.084 |
| LSTM | 4-5-1 | 0.013 |

IV. BEST NETWORK MODEL FOR EUR/NPR

| Model | Best Structure | MAE |
|---|---|---|
| MLP | 4-6-1 | 0.12 |
| SRNN | 4-4-1 | 0.18 |
| GRU | 4-7-1 | 0.064 |
| LSTM | 4-5-1 | 0.042 |

From the above table, it clarifies that LSTM model of each currency produce the lowest prediction error. Also, the model that best fits for one currency may not fit for other currency to that extent. Hence, the number of hidden neurons for each model to operate with different currency differs from one another.

V. BEST NETWORK MODEL FOR GBP/NPR

| Model | Best Structure | MAE |
|---|---|---|
| MLP | 4-5-1 | 0.052 |
| SRNN | 4-4-1 | 0.214 |
| GRU | 4-7-1 | 0.0177 |

| LSTM | 4-5-1 | 0.0388 |

On the basis of above analysis, for USD/NPR using LSTM, 4-5-1 neural network was observed to be the best structure likewise EUR/NPR was best modeled by 4-5-1 neural network and GBP/NPR was best modeled by 4-7-1 neural network.

From 36 recurrent neural network models of USD/NPR, LSTM model number 4 with 5 hidden neurons has the lowest forecast error (MAE - 0.013). Similarly, for EUR/NPR, LSTM model number 4 with 5 hidden neurons showed the best results with MAE - 0.042 and for GBP/NPR, GRU model number 6 with 7 hidden neurons showed the best results with MAE-0.0177. Among all different 106 models, LSTM model number 4 with 5 hidden neurons have the minimum MAE. Among all the 81 recurrent network models, LSTM model number 4 was concluded as the best model for prediction among RNN architecture. Figure 4, 5 and 6 show the graph of actual vs predicted value of different currencies according to models.

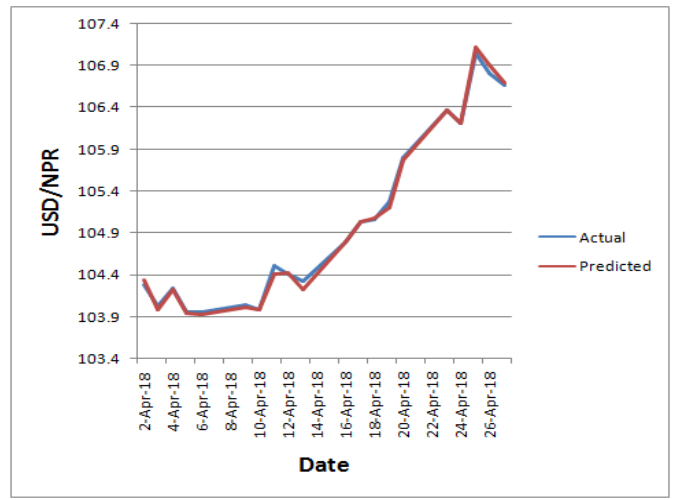

4. Actual VS Prediced Value Graph Of USD/NPR

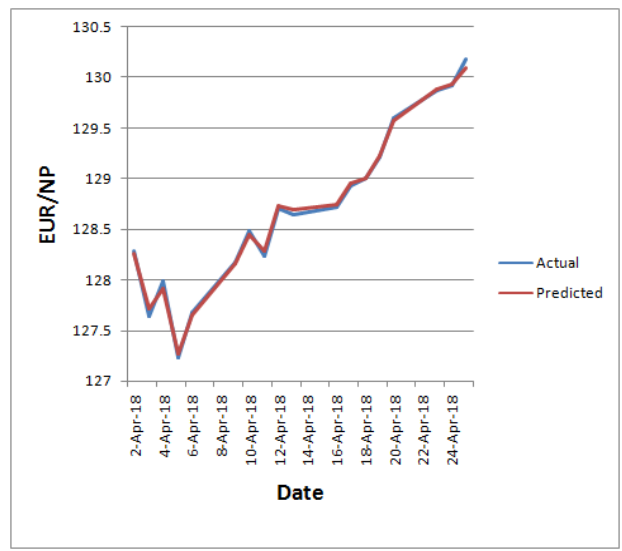

5. Actual VS Prediced Value Graph Of EUR/NPR

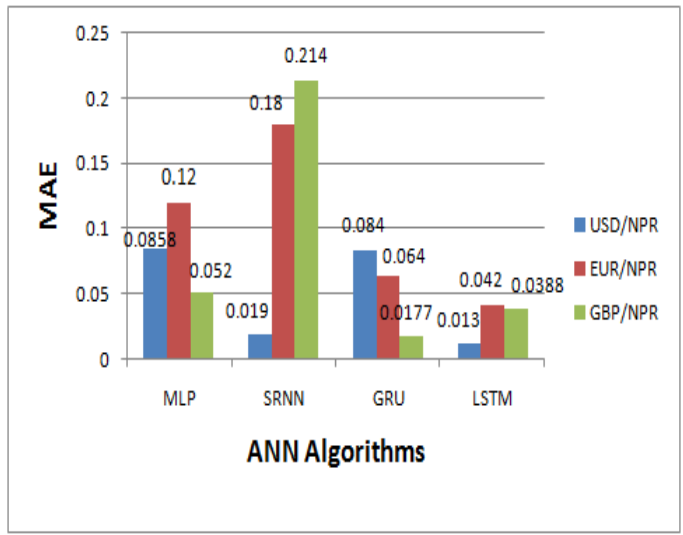

3. Showing MAE For Different Algorithm Of Each Currency

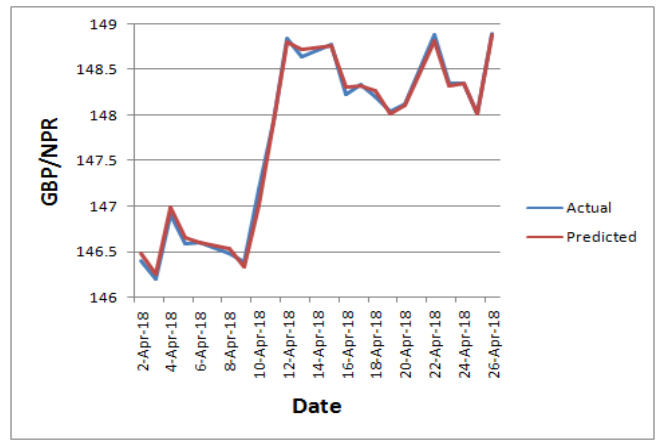

6. Actual VS Prediced Value Graph Of GBP/NPR

## VI. Conclusion And Future Enhancement

This paper investigates prediction of Nepalese currency against American Dollar, Euro and Pound Sterling using different ANN model. Results demonstrate that LSTM model can predict the foreign exchange rates and have better performance than the other models. Different trials were performed by changing the number of hidden neurons until the best result was found. After several experiments with different network architectures, LSTM with structure 4-5-1 gave the most accurate results in terms of MAE. We would like to add more parameters on input for more accurate prediction and also minimize time, space complexity for processing ANN architecture.